\pdfoutput=1
\documentclass[runningheads]{llncs}

\usepackage[T1]{fontenc}
\usepackage[utf8]{inputenc}
\usepackage{graphicx}
\usepackage{amsmath}
\usepackage{amssymb}
\usepackage{algorithm}
\usepackage{algorithmic}
\usepackage{tabularx}
\usepackage{array}
\usepackage{booktabs}
\usepackage{subcaption}
\usepackage{hyperref}
\usepackage{color}

\begin{document}

\title{Recovering Input Text from Hidden States:\\ Study of Gradient-Based Inversion of \\ Decoder-Only Language Models}

\titlerunning{Gradient-Based Inversion of Decoder-Only Language Models}

\author{Miko\l{}aj S\l{}owikowski\inst{1} \and Maciej Witold Majewski\inst{1}}

\authorrunning{M. S\l{}owikowski and M.W. Majewski}

\institute{AGH University of Krakow, Faculty of Physics and Applied Computer Science,\\
al. Mickiewicza 30, 30-059 Krakow, Poland\\
\email{mmajewski@agh.edu.pl, mikslo@student.agh.edu.pl}}

\maketitle

\begin{abstract}
This work studies
the \emph{hidden-state inversion problem}: recovering the original input token
sequence of a decoder-only language model from its last-layer hidden states.
Rather than treating inversion as a one-shot reconstruction, we study it as a
\emph{continuous embedding-space optimisation} in which a soft proxy is driven
towards the leaked target without any hard-token projection during the search,
and a token is committed only once, at the end of the inner loop. This design
choice has two consequences which are the main focus of this paper. First, keeping
the optimisation entirely in continuous space exposes a rich set of internal
signals---rank trajectories of the ground-truth token, per-position loss curves,
and a discrete loss measured at commit time. Second, the discrete loss yields allows to asses the correctness of recovery via cumulative discrete loss. We further analyse \emph{which} tokens break the
reconstructions and find a sharp categorical asymmetry: space-prefixed, high-frequency
function words in dense regions of the embedding matrix dominate the failures,
while content-bearing tokens are recovered almost perfectly. On 10-token C4
prompts the exact-match rate rises from 66.9\% to 97.5\% (mean similarity 0.994)
as the candidate window is widened, confirming that most errors are recoverable
near-misses rather than genuine ambiguities. A comparison with the released
SIPIT~\cite{nikolaou2025languagemodelsinjectiveinvertible} reference situates these findings: per-step hard projection is faster, but
the continuous formulation is what makes the optimisation observable and its
failures detectable. The results show that last-layer hidden states of GPT-2 are
as sensitive as the original text.

\end{abstract}

\section{Introduction}
\label{sec:introduction}

Large language models (LLMs) have become central infrastructure for a growing
range of applications, and their deployment increasingly moves beyond a
single-machine setting. A client device may run the early layers of a model
locally and send only the resulting activations to a remote server that
completes the computation; participants in a federated pipeline exchange
gradient updates that carry information about their private inputs; retrieval
systems precompute and cache hidden-state representations of sensitive
documents for later reuse. In each of these scenarios raw text never leaves the
trusted perimeter---only tensors of floating-point numbers do.

This separation between text and its neural representation is typically treated
as a privacy boundary, under the assumption that a hidden-state vector is an
opaque, high-dimensional encoding from which the original words cannot be
recovered. This assumption underpins split-inference
protocols~\cite{vepakomma2018split}, embedding APIs that return dense vectors
without the original query~\cite{morris2023vec2text}, and federated
fine-tuning that transmits gradient tensors carrying implicit information about
training examples~\cite{mcmahan2017federated,zhu2019dlg}. If the assumption of
irreversibility holds, an adversary who intercepts the tensors learns nothing
about the text; if it does not, the entire privacy argument collapses.

Recent theoretical work shows that the assumption is in fact incorrect for
decoder-only transformer models under mild conditions. The SIPIT
result~\cite{nikolaou2025languagemodelsinjectiveinvertible} establishes that the
mapping from a token sequence to its last-layer hidden states is
\emph{injective}: no two distinct input sequences produce the same sequence of
hidden-state vectors. Injectivity implies that the hidden states carry exactly
the same information as the original tokens, and that in principle a
sufficiently powerful algorithm could recover the input exactly. The practical
question is therefore not \emph{whether} inversion is possible in theory, but
\emph{how efficiently} it can be carried out and under what conditions it
succeeds.

This paper addresses that practical question directly with a gradient-based
approach: a continuous proxy vector in embedding space is optimised, position by
position, so that the model's predicted hidden state matches the leaked target;
once the proxy has converged it is projected onto the nearest discrete token
embedding and the recovered token is committed. The method requires white-box
access to the model---its weights, tokenizer, and embedding matrix---and no
knowledge of the original text beyond the shape of the hidden-state tensor.

\subsection{Problem Statement}
\label{subsec:problem}

Let \(\mathcal{V}\) be the vocabulary of a fixed, pretrained decoder-only
language model \(\mathcal{M}\) with \(L\) transformer layers and hidden
dimension \(n_\mathrm{emb}\). Take an input sequence of \(T\) tokens
\(x = (x_1,\dots,x_T)\in\mathcal{V}^T\). A single forward pass produces one
last-layer hidden-state vector \(h_t(x)\in\mathbb{R}^{n_\mathrm{emb}}\) per
position \(t\). We collect these \(T\) vectors as the rows of a matrix
\(H(x)\in\mathbb{R}^{T\times n_\mathrm{emb}}\), so that row \(t\) is \(h_t(x)\).
Because \(\mathcal{M}\) uses causal (left-to-right)
self-attention~\cite{vaswani2017attention} exclusively, \(h_t(x)\) depends only
on the prefix \((x_1,\dots,x_t)\); the rows are therefore well-defined from that
single forward pass. The map \(x\mapsto H(x)\) is the \emph{forward} direction;
this paper studies its \emph{inverse}---recovering the input sequence \(x\) from
the matrix \(H(x)\) alone.

\subsection{Injectivity and Solvability} 
\label{subsec:injectivity}

The problem is well-posed only if \(x\mapsto H(x)\) is injective, i.e.\ no two
distinct sequences produce the same hidden-state matrix. Nikolaou et
al.~\cite{nikolaou2025languagemodelsinjectiveinvertible} prove that this mapping
is injective for transformer language models under some conditions, all of
which hold for GPT-2. Injectivity means that the hidden states pin down the
input: each matrix \(H(x)\) comes from exactly one sequence \(x\). Recovery
therefore has a single correct answer; the difficulty lies in computing it
efficiently.

\subsection{Related Work}
\label{subsec:related}

\paragraph{Embedding inversion.}
Song and Raghunathan show that embedding models expose sensitive attributes of
the underlying text~\cite{song2020embeddingleakage}. Morris et al.\
(\textsc{vec2text}) recover inputs from sentence embeddings by training a
corrector model that iteratively refines a hypothesis to match the target
embedding~\cite{morris2023vec2text}, and Li et al.\ (\textsc{geia}) decode
coherent text directly from a single pooled embedding~\cite{li2023geia}; broader
analyses confirm that such leakage is widespread~\cite{pan2020privacy}. Unlike
these methods, which operate on a single pooled vector and typically train a
dedicated inversion model, we target a sequence of per-token last-layer hidden
states---a richer, causally structured signal---which lets the problem be
decomposed position by position and evaluated by exact token-level
reconstruction rather than semantic similarity.

\paragraph{Privacy attacks on split and federated inference.}
The stakes are highest in distributed architectures that exchange intermediate
representations: split inference crosses the network boundary with an activation
tensor~\cite{vepakomma2018split}, and federated learning shares gradient updates
rather than raw examples~\cite{mcmahan2017federated}. A closely related line
attacks the exchanged \emph{gradients}: Zhu et al.\ reconstruct training
examples (\emph{Deep Leakage from Gradients})~\cite{zhu2019dlg}, \textsc{tag}
recovers input text from gradients~\cite{deng2021tag}, and \textsc{lamp} couples
gradient matching with a language-model prior~\cite{balunovic2022lamp}.
Hidden-state inversion is complementary: it targets forward activations crossing
a split-inference boundary rather than backward gradients, but shares the
message that intermediate tensors leak their inputs.

\section{Method}
\label{sec:method}

The algorithm has three layers: an outer loop over token positions, an inner
gradient-based loop that optimises a continuous proxy for the current position,
and a discrete verification stage that turns the optimised proxy into a
committed token. Algorithm~\ref{alg:inversion} states the procedure. The design
principle that distinguishes it from per-step projection schemes is a strict
\emph{separation between the continuous search and the discrete decision}: the
proxy is never snapped onto a real token during optimisation, and a token id is
produced exactly once, after the inner loop has converged. This separation is
not merely a stylistic choice---it is what makes the optimisation observable
(Section~\ref{sec:results}), because the gap between the converged continuous
proxy and the committed discrete token can be measured directly and used as an
error signal.

\begin{algorithm}[t]
\caption{Token-wise hidden-state inversion}
\label{alg:inversion}
\begin{algorithmic}[1]
\REQUIRE target hidden states \(H=(h_1,\dots,h_T)\), model \(f\), embedding
matrix \(E\), thresholds
\STATE \(\texttt{recovered}\leftarrow[\,]\)
\FOR{\(t=1\) to \(T\)}
    \STATE initialise proxy embedding \(e_t\) (embedding of a random token id)
    \FOR{optimisation step \(k=1,\dots,K\)}
        \STATE predict \(\hat{h}_t=f(e_t \mid \texttt{recovered})\) conditioned on the committed prefix
        \STATE compute loss \(\mathcal{L}_t=\mathrm{MSE}(\hat{h}_t,h_t)\)
        \STATE update \(e_t\) with Adam (cosine-annealed LR, gradient clipping)
        \IF{\(\mathcal{L}_t<\texttt{loss\_th}\)} \STATE \textbf{break} \ENDIF
    \ENDFOR
    \STATE retrieve top-\(C\) nearest candidates in \(E\) to \(e_t\)
    \STATE verify candidates by a forward pass; pick first with discrete error
    below threshold
    \STATE if none passes, fall back to the top-1 nearest candidate
    \STATE commit token id and append it to \texttt{recovered}
\ENDFOR
\RETURN recovered token ids
\end{algorithmic}
\end{algorithm}

\paragraph{Inner optimisation.}
At each position the proxy \(e_t\in\mathbb{R}^{n_\mathrm{emb}}\) is optimised by
Adam~\cite{kingma2015adam} to minimise the MSE between the predicted and target
hidden state. The learning rate follows a cosine-annealing schedule with
\(T_\mathrm{max}=K\), and gradients are clipped to a maximum norm of
\(1.0\)~\cite{pascanu2013clipping}; together these tame the early-step
instability of Adam on this strongly non-convex objective. The loop exits early
when \(\mathcal{L}_t\) drops below \texttt{loss\_th}, otherwise it runs the full
budget \(K\). Most tokens are solved well under \(K\) steps; the full budget is
consumed only for hard tokens, such as rare sub-word pieces or tokens whose
embedding neighbourhood is densely populated.

\paragraph{Proxy initialisation.}
The proxy is initialised to the embedding of a uniformly random token id, so
optimisation starts on the discrete embedding manifold rather than at an
arbitrary point; this consistently reduces the steps needed to reach the loss
threshold. Zeros and a random Gaussian vector are evaluated as alternatives in
Section~\ref{sec:results}.

\paragraph{Discrete verification and commit.}
After the inner loop exits, the top-\(C\) nearest tokens to \(e_t\) (under the
\(\ell_1\) metric) are retrieved and tested in distance order with one discrete
forward pass each; the first candidate that reproduces \(h_t\) within tolerance
is committed. If none passes, the nearest candidate is committed unconditionally
as a fallback, with no guarantee that it reproduces the target. This fallback is
rare in practice but unavoidable without global search. Crucially the pipeline
performs exactly \emph{one} discrete search per position---only at the end of
the inner loop---in contrast with per-step projection strategies
(Section~\ref{subsec:ref-comp}). Committing once means that the discrete loss
\(\mathcal{L}_t^{\mathrm{disc}}\), the MSE obtained when the committed token id
is fed forward, is a clean, well-defined quantity at every position: it measures
how faithfully the \emph{discrete} decision reproduces the target, independently
of how far the continuous proxy converged. Section~\ref{subsec:diagnostics}
turns this quantity into an online indicator of whether the inversion has gone
wrong.

\section{Experimental Setup}
\label{sec:setup}

Two English corpora are used. The \texttt{allenai/c4}
corpus~\cite{raffel2020c4}, a large-scale web crawl covering many domains, is the
primary batch benchmark: for each sampled document the first 10 byte-pair
(BPE)~\cite{sennrich2016bpe} tokens (no special tokens) form the inversion
target. The
\texttt{wikimedia/wikipedia} snapshot (\texttt{20231101.en}) is used for
loss-curve and rank-trajectory analyses; sentences of 25--70 characters are
extracted, tokenized, and truncated to at most 10 tokens.

Development and most experiments ran on a workstation with NVIDIA RTX A6000 GPUs
(48~GB, CUDA~12.5); GPT-2 small fits in a single GPU, so every run uses one GPU.
Larger batches were submitted to the Athena cluster (ACK Cyfronet AGH) via SLURM
with full-precision inference matching the workstation.

We report \emph{exact match} (EM), the fraction of prompts recovered correctly
at \emph{every} position; \emph{token accuracy}, the fraction of correct
positions; and character-level \emph{SequenceMatcher similarity}
(\texttt{difflib}), which gives partial credit for near-correct reconstructions.
For diagnostics we also record the per-step continuous loss, the
\emph{discrete loss} \(\mathcal{L}_t^{\mathrm{disc}}\) (MSE when the committed
token id is fed forward), and the \emph{rank} of a token \(v^\star\) with
respect to a proxy \(u\), \(\mathrm{rank}(v^\star;u)=|\{v:\|E[v]-u\|<\|E[v^\star]-u\|\}|\),
i.e.\ the number of vocabulary tokens strictly closer to \(u\) than \(v^\star\)
(rank 0 = nearest neighbour). Per-token wall time is the time to recover a
single position, from proxy initialisation through the inner optimisation loop
to committing the token id.

The Baseline approach uses Adam with initial learning rate \(\alpha_0=0.05\), cosine
annealing (\(T_\mathrm{max}=K\)), weight decay \(10^{-5}\), gradient-clip norm
\(1.0\), step budget \(K=1000\), stopping threshold
\(\texttt{loss\_th}=10^{-4}\), \(\ell_1\) candidate metric, max candidates
\(C=2000\), 10-token prompts, and fixed seed 42. Each ablation varies exactly
one parameter while holding the rest fixed.

\section{Results}
\label{sec:results}

During the studies we perform multiple hyperparameter optimisiations, as the method relies heavy on gradient optimisation.

\begin{figure}[t]
\centering
\includegraphics[width=0.92\textwidth]{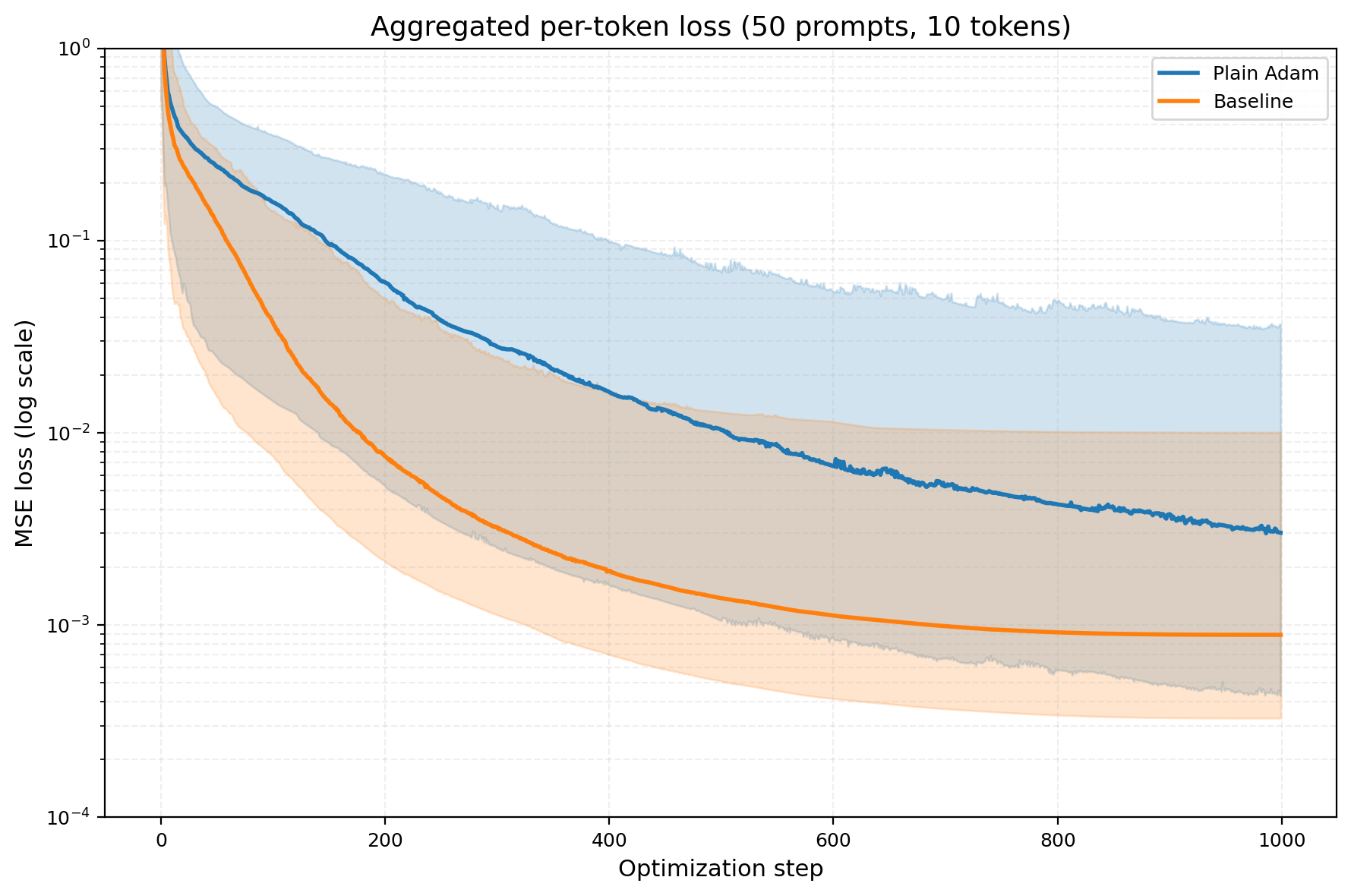}
\caption{Per-token continuous loss curves under Adam versus Adam with
cosine annealing and gradient clipping (baseline). Without stabilisation the loss spikes
early and the trajectory is erratic; with it the descent is smooth and monotone.}
\label{fig:adamloss}
\end{figure}

\begin{figure}[t]
\centering
\includegraphics[width=0.92\textwidth]{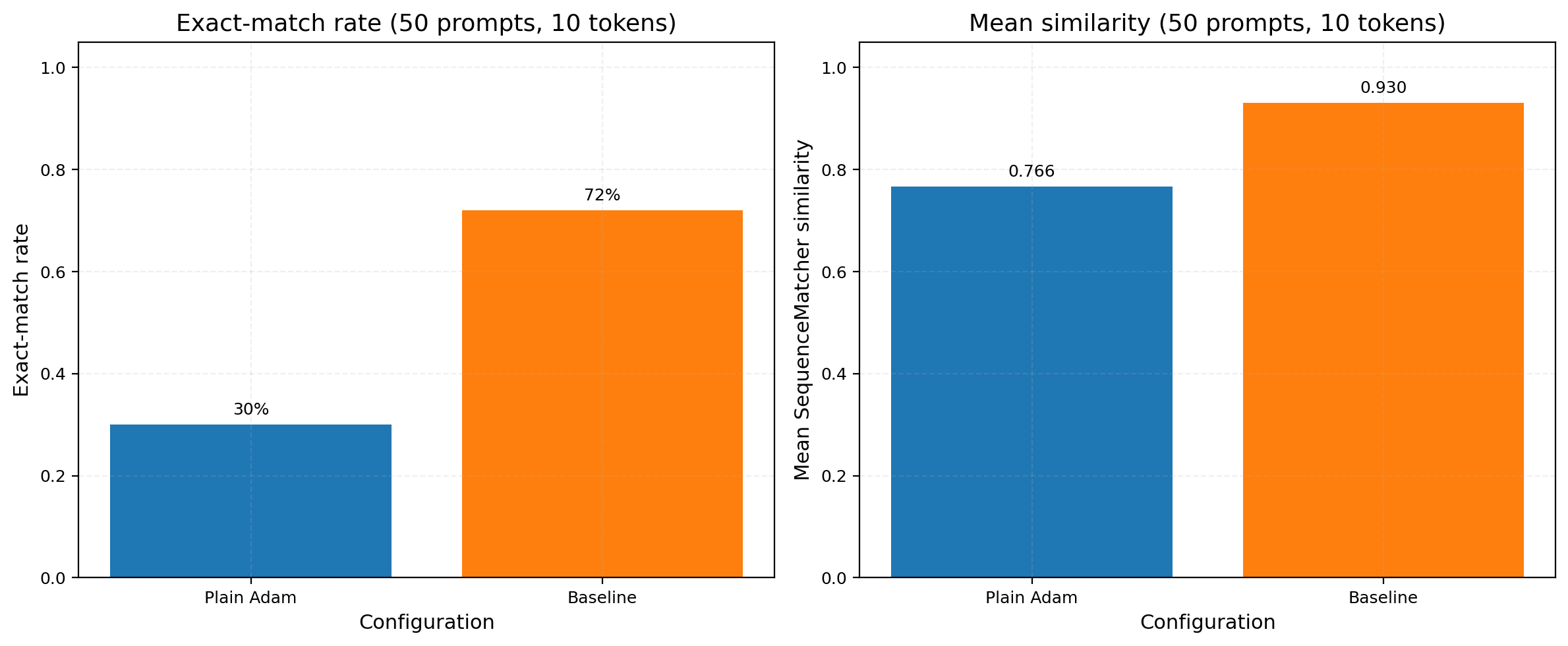}
\caption{Adam stabilisation: effect of cosine annealing and gradient clipping on
exact-match rate and mean similarity. Plain Adam suffers an early loss spike;
adding the schedule and clipping removes it and lifts both metrics.}
\label{fig:adam}
\end{figure}

Plain Adam on this objective is unstable: an early loss spike can derail the
trajectory before it settles, as the raw loss curves in
Fig.~\ref{fig:adamloss} show. Adding cosine annealing and gradient clipping
removes the spike, yields a smooth monotone descent, and improves both exact
match and similarity (Fig.~\ref{fig:adam}); this stabilised configuration is used
everywhere below.

A learning rate sweep over \(\alpha_0\in\{0.01,\dots,0.5\}\) identifies \(\alpha_0=0.05\) as the
best trade-off between convergence speed and stability; larger rates converge
faster initially but overshoot the narrow basin around the true embedding.

\begin{figure}[t]
\centering
\includegraphics[width=0.92\textwidth]{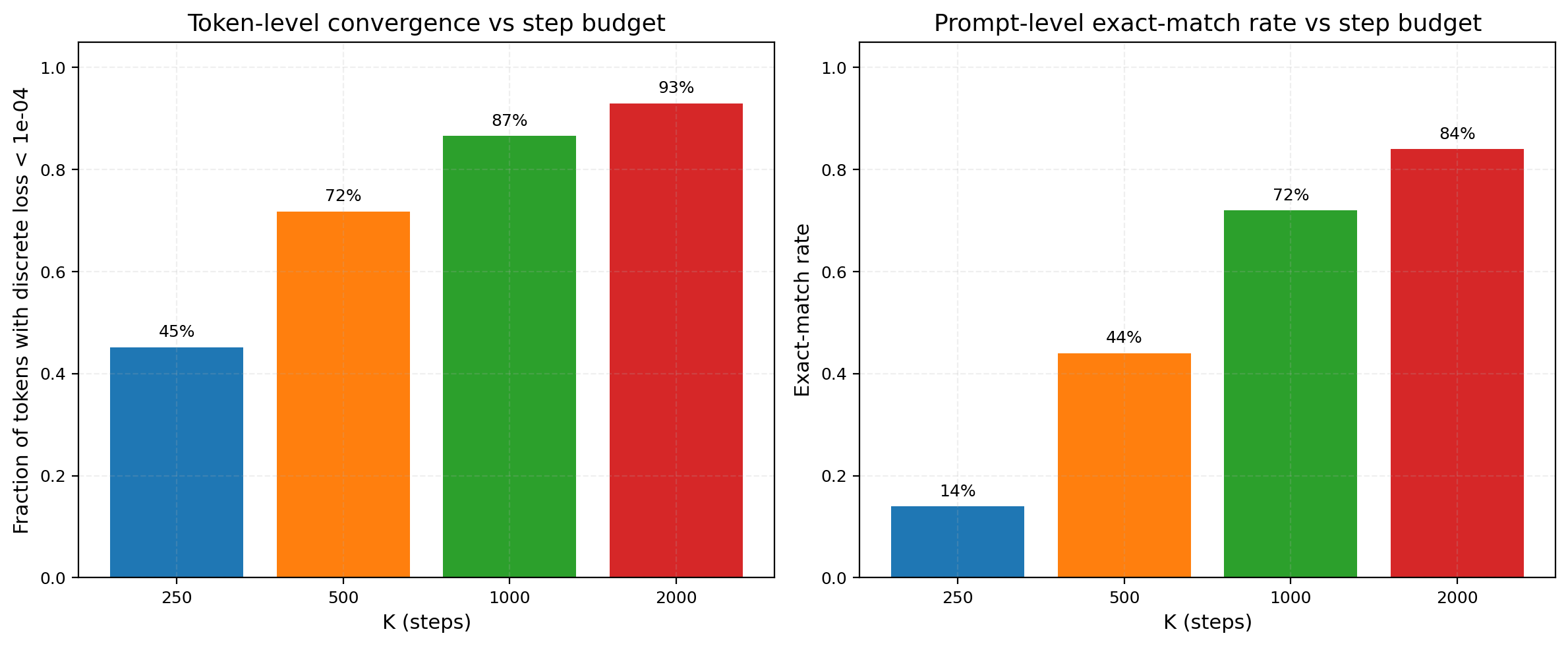}
\caption{Exact-match rate vs.\ step budget \(K\) (\(C=2000\), C4 prompts).
Accuracy rises steeply up to \(K\approx1000\), after which returns diminish.}
\label{fig:ksweep}
\end{figure}

Exact match grows steeply with the step budget \(K\) ---from roughly 14\% at
\(K=250\) to 84\% at \(K=2000\) (with \(C=2000\))---as shown in
Fig.~\ref{fig:ksweep}. Returns diminish past \(K\approx1000\), which motivates
the Baseline operating point.

\begin{figure}[t]
\centering
\includegraphics[width=0.8\textwidth]{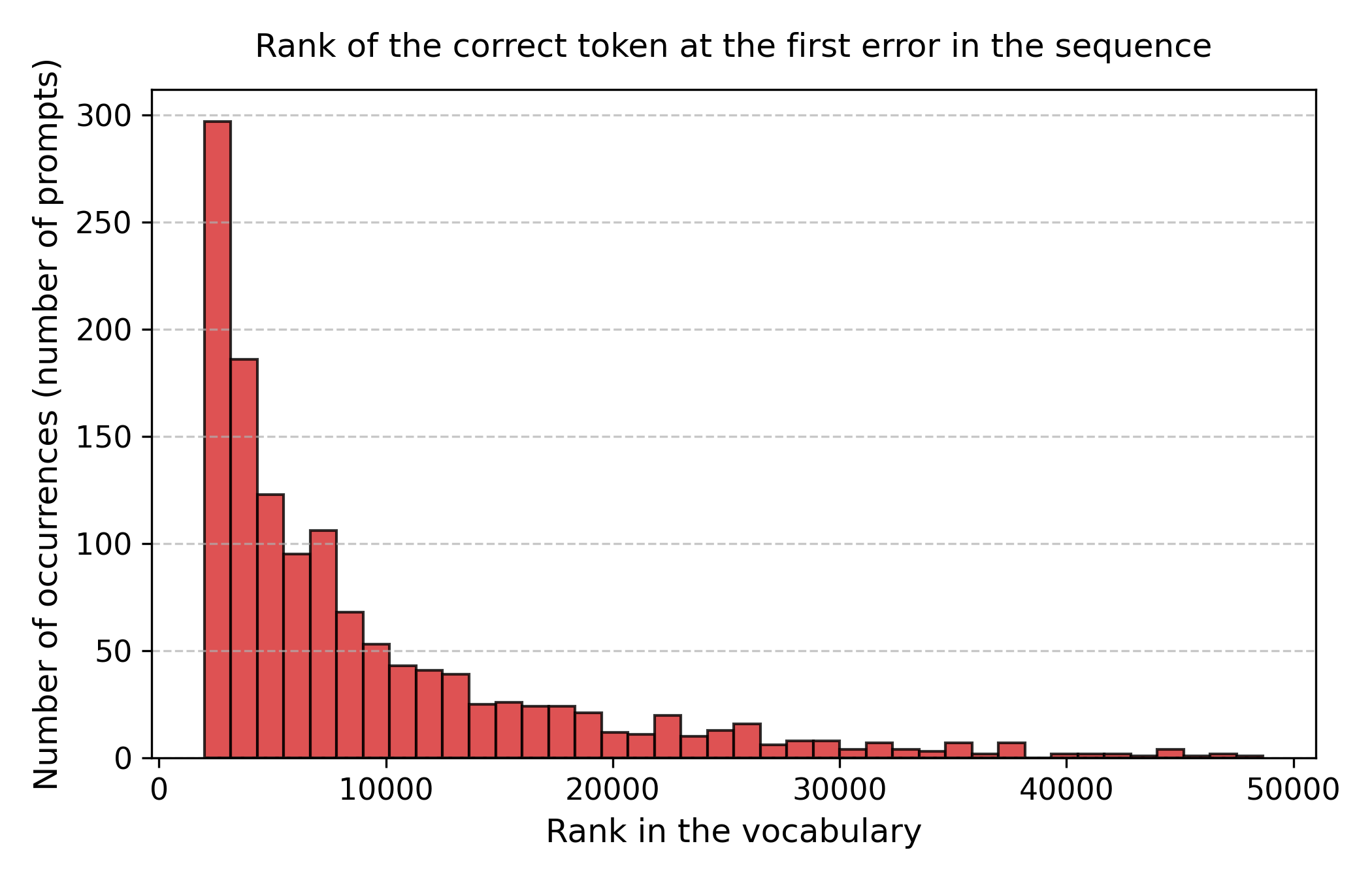}
\caption{Rank of the ground-truth token at the first wrongly committed position
under the \(\ell_1\) metric, aggregated over 2000 prompts. The mass is
heavily concentrated near rank~0: even at error positions the correct token is
rarely far from the top of the \(\ell_1\)-sorted candidate list, which is why a
wider candidate window \(C\) recovers most near-misses.}
\label{fig:l1hist}
\end{figure}

A stopping threshold of \(10^{-4}\) cleanly separates converged from non-converged
positions without wasting steps. Among \(\ell_1\), cosine, and dot-product
candidate metrics, \(\ell_1\) keeps the ground-truth token closest to the top of
the sorted candidate list at error positions; we therefore adopt \(\ell_1\) as
the default. Fig.~\ref{fig:l1hist} confirms this on a 2000-position sample: even
when the algorithm commits the wrong token, the correct token is rarely far from
rank~0 under \(\ell_1\).

Random-token and zero initialisation both start near or on the embedding
manifold and outperform a random Gaussian start; the two are close, with zeros
used for the final large-batch operating points.

\subsection{Rank Evolution and Operating Points}
\label{subsec:rank}

\begin{figure}[t]
\centering
\includegraphics[width=0.9\textwidth]{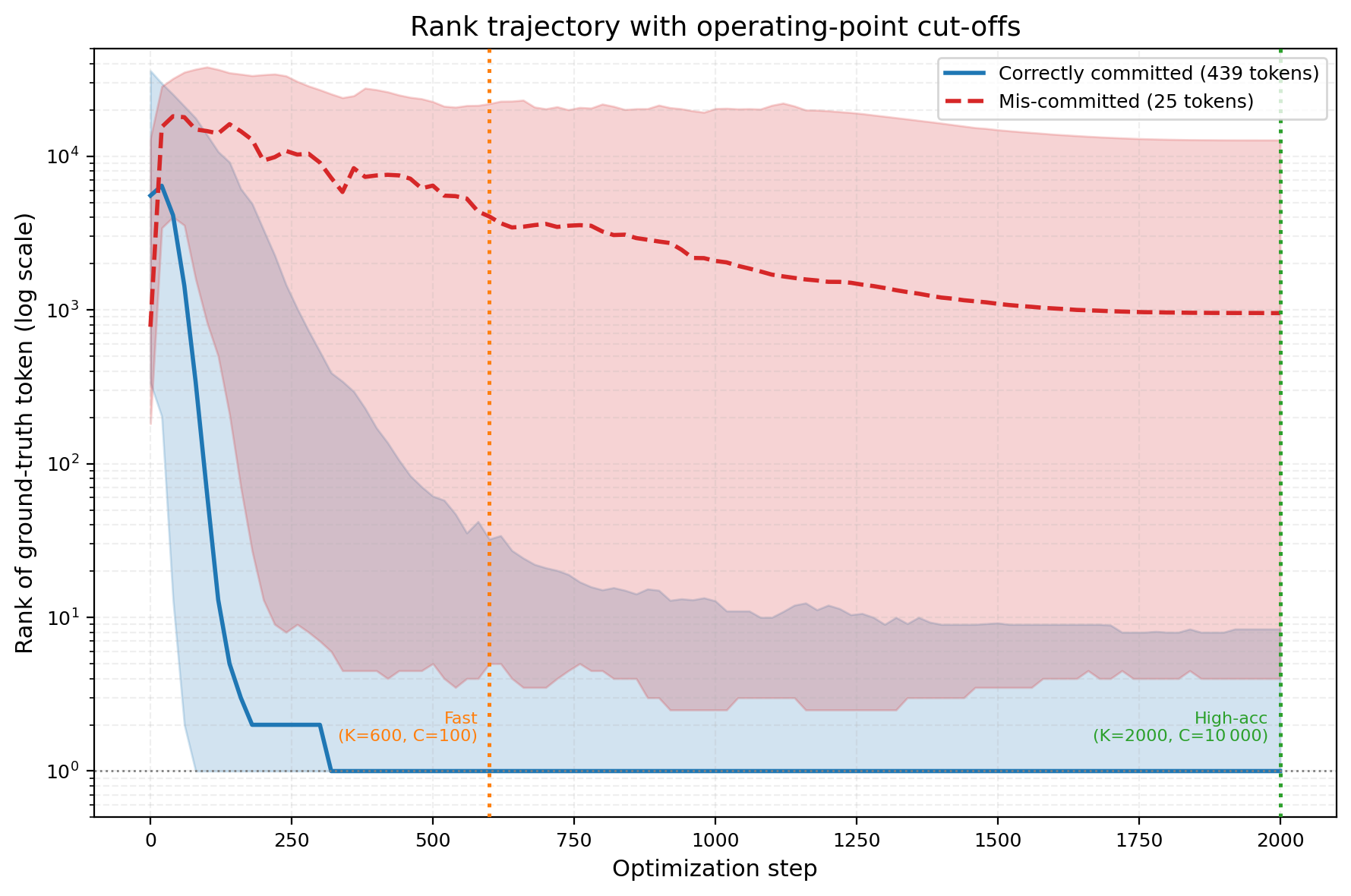}
\caption{Rank trajectory of the ground-truth token for the \(K=2000\) run
(\(N=50\) C4 prompts, 10 tokens each). Blue: correctly committed tokens (443);
red dashed: mis-committed tokens (26); shaded bands are 10--90th percentile. The
orange dotted line marks step~600 (Fast, \(C=100\)); the green dotted line marks
step~2000 (High-accuracy, \(C=10\,000\)).}
\label{fig:rankevo}
\end{figure}

The rank-evolution trajectories recorded during the step-budget ablation provide
the principal evidence for the final operating points
(Fig.~\ref{fig:rankevo}). For \emph{correctly} committed tokens the median rank
of the ground-truth token descends to rank~1 well before step~300: easy tokens
have already converged by the time the Fast configuration terminates at
step~600, so the Fast cut-off sacrifices no accuracy on them. For
\emph{mis-committed} tokens the median never approaches rank~1 regardless of step
count: the optimiser fails to reach the correct neighbourhood for this minority,
and the only remedy is a \emph{wider candidate window}, not more steps. This
directly motivates the three operating points in
Table~\ref{tab:operating}: Fast (\(K=600\), \(C=100\)), Baseline (\(K=1000\),
\(C=2000\)), and High-accuracy (\(K=2000\), \(C=10\,000\)).

\begin{table}[t]
\centering
\begin{tabular}{lrr}
\toprule
\textbf{Configuration} & \boldmath$K$ & \boldmath$C$ \\
\midrule
Fast        & 600  & 100 \\
Baseline    & 1000 & 2000 \\
High-accuracy & 2000 & 10\,000 \\
\bottomrule

\end{tabular}
\caption{Operating-point configurations. The fixed parameters (Adam,
\(\alpha_0=0.05\), cosine schedule, clip norm 1.0, \(\texttt{loss\_th}=10^{-4}\),
\(\ell_1\)) are shared; the table varies the step budget \(K\) and candidate
window \(C\).}
\label{tab:operating}
\end{table}

\subsection{Inversion Accuracy}
\label{subsec:accuracy}

Table~\ref{tab:accuracy} reports exact match and mean similarity across the three
configurations and the two datasets. With C4 dataset the exact-match rate rises from
66.9\% (Baseline, \(N=450\)) to 97.5\% with mean similarity 0.994
(High-accuracy, \(N=200\)). The interpretation is reinforced by
Section~\ref{subsec:rank}: increasing \(C\) converts near-misses into successes
without touching the inner loop. The Fast configuration is markedly harder on
Wikipedia (10\%) than on C4 (35\%), reflecting Wikipedia's more uniform
vocabulary and shorter, less formulaic sentences; a fully controlled
cross-dataset comparison with identical hyperparameters remains future work.

\begin{table}[t]
\centering

\begin{tabular}{llrrlrrrr}
\toprule
\textbf{Config} & \textbf{Dataset} & \boldmath$K$ & \boldmath$C$ & \textbf{init}
& \boldmath$T$ & \boldmath$N$ & \textbf{EM(\%)} & \boldmath$SM_{mean}$ \\
\midrule
Fast        & C4        & 600  & 100   & zeros & 10 & 100 & 35.0          & 0.800 \\
Fast        & Wikipedia & 600  & 100   & zeros & 10 & 100 & 10.0          & 0.711 \\
Baseline    & C4        & 1000 & 2000  & zeros & 10 & 450 & \textbf{66.9} & 0.918 \\
High-acc    & C4        & 2000 & 10\,000 & zeros & 10 & 200 & \textbf{97.5} & \textbf{0.994} \\
\bottomrule
\end{tabular}
\caption{Inversion accuracy across configurations and datasets. \(T\) = prompt length; \(N\) =
number of evaluation samples; init = proxy
initialisation.}
\label{tab:accuracy}
\end{table}

\subsection{Self-Diagnostic Failure Detection}
\label{subsec:diagnostics}

\begin{figure}[t]
\centering
\includegraphics[width=0.92\textwidth]{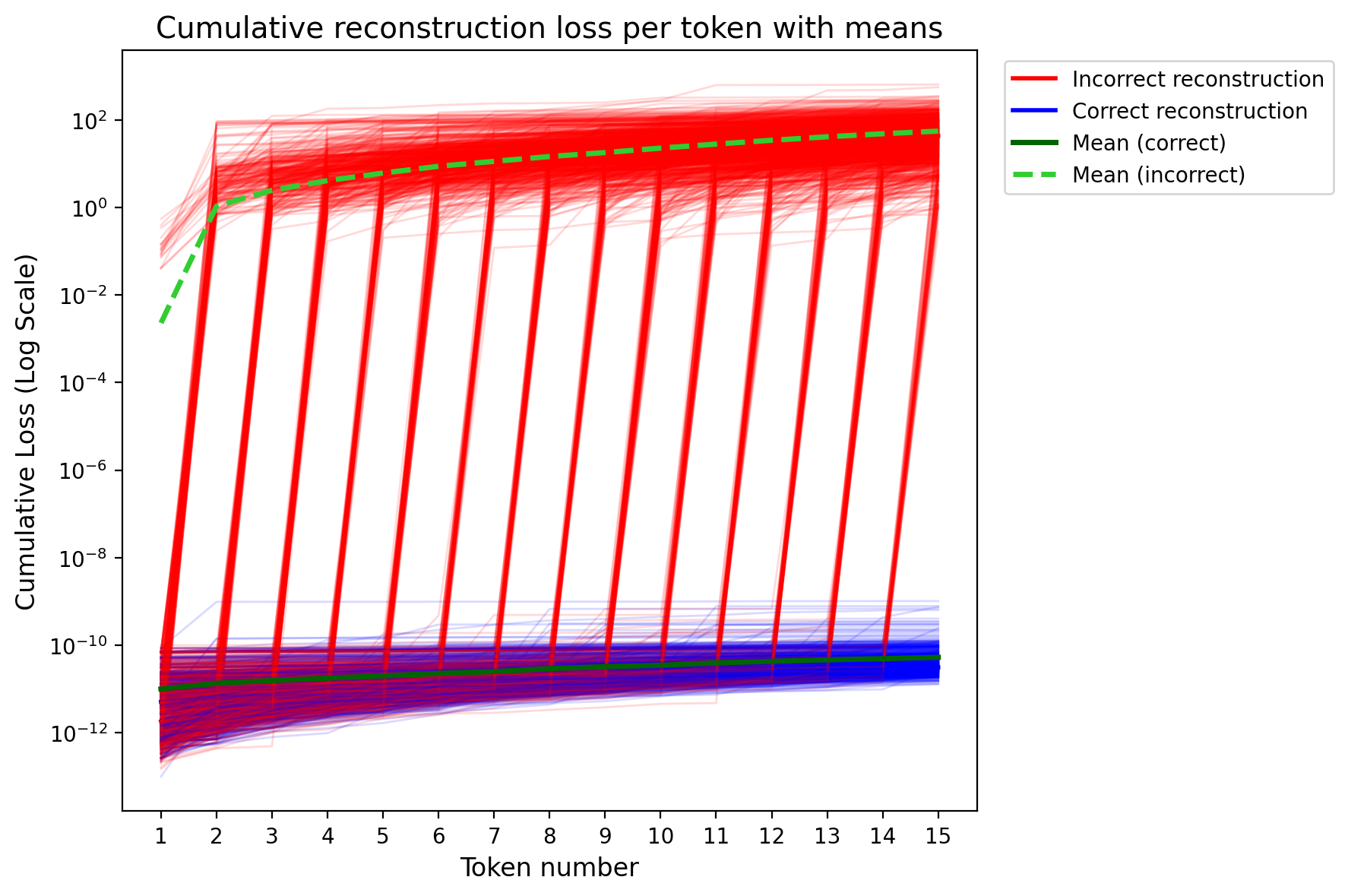}
\caption{Cumulative discrete loss per token position (2000-prompt batch,
logarithmic \(y\)-axis). Blue: prompts recovered exactly; red: prompts with at
least one error; the green lines are the per-position means over the correct and
incorrect subsets. The two regimes are separated by roughly ten orders of
magnitude, which is what makes the discrete loss usable as a ground-truth-free
failure detector.}
\label{fig:cumloss}
\end{figure}

The single-commit design (Section~\ref{sec:method}) yields a clean discrete loss
\(\mathcal{L}_t^{\mathrm{disc}}\) at every position, and this turns out to be the
most practically useful by-product of keeping the optimisation in continuous
space. Plotting the cumulative discrete loss
\(\sum_{i\le t}\mathcal{L}_i^{\mathrm{disc}}\) against position
(Fig.~\ref{fig:cumloss}) reveals a sharp, binary signal. Prompts recovered
exactly produce cumulative curves that hug the floor of the plot
(\(\sim 10^{-11}\)) for the entire sequence, because each committed token
reproduces its target hidden state to floating-point precision. Prompts with at
least one error show a pronounced step at the position of the first wrong
commitment, followed by a steep rise as the corrupted recovery propagates the
error to every later position. On the logarithmic axis the two regimes are
separated by roughly \emph{ten orders of magnitude} by the end of the sequence.

We draw two consequences from this property. First, the cumulative discrete loss
serves as a self-diagnostic signal: a spike marks the position of a reconstruction
error, so the algorithm can flag a failed recovery online without access to the
original sequence. Second, the existence of positions that converge in continuous
space yet commit the wrong token indicates that the embedding-space objective
contains flat or near-degenerate regions in which the proxy settles away from the
correct discrete neighbour. This points to a route for improving efficiency:
better-conditioned search spaces or alternative search strategies that escape such
regions may raise accuracy without enlarging the candidate window.

\begin{figure}[t]
\centering
\includegraphics[width=0.92\textwidth]{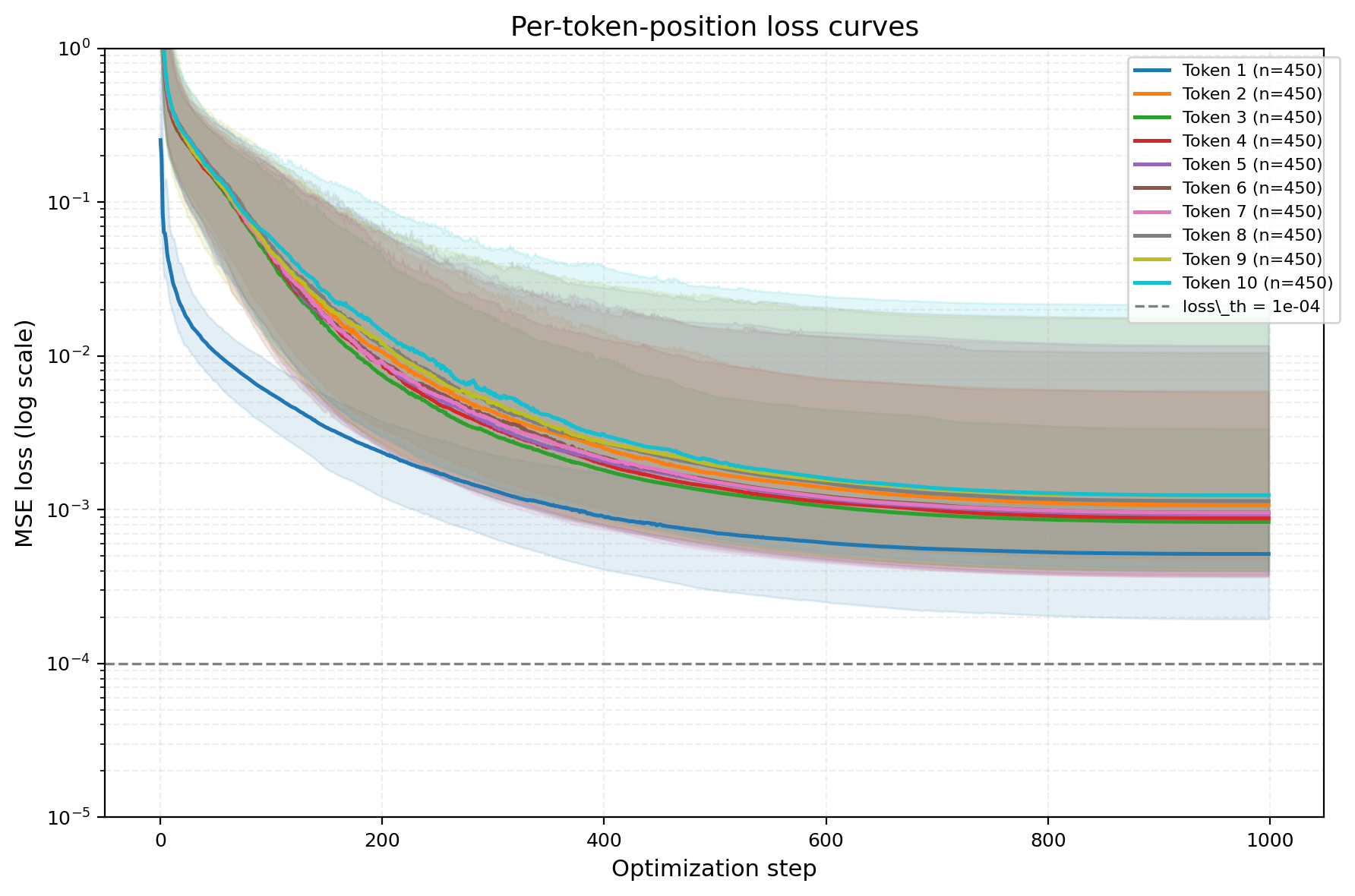}
\caption{Median continuous loss vs.\ optimisation step, grouped by token
position within the 10-token prompt (\(K=1000\), \(N=450\) C4 prompts). Shaded
bands are the 10--90th percentile; the dashed line marks the stopping threshold
\(\texttt{loss\_th}=10^{-4}\). The first position converges fastest (its
optimisation is unconditioned) and later positions are progressively harder.}
\label{fig:losspos}
\end{figure}

Additionally, the per-position loss curves (Fig.~\ref{fig:losspos}) explain \emph{why}
difficulty grows along the sequence. The first token converges fastest---its
optimisation is unconditioned (no committed prefix), so the landscape is
comparatively smooth---while the median curve rises monotonically with position,
consistent with a longer committed prefix constraining the search more tightly
and compounding any residual mismatch from earlier positions. Across all
positions the median plateaus above \(\texttt{loss\_th}\), reflecting the mix of
easy tokens that exit early and hard tokens that exhaust the budget; the latter
are dominated by the function-word category analysed in
Section~\ref{subsec:problematic}.

\subsection{Runtime}
\label{subsec:runtime}

Per-token times are bimodal: tokens that trigger early exit cluster at a low-time
mode, while tokens that exhaust the budget peak near \(K\) times the per-step
cost. Table~\ref{tab:runtime} reports mean times per token and per 10-token
prompt on a single RTX A6000. Wall time scales with the step budget, from
9.7~s/token (Fast) to 27.5~s/token (High-accuracy).

\begin{table}[t]
\centering

\begin{tabular}{llrrrr}
\toprule
\textbf{Config} & \textbf{Dataset} & \boldmath$K$ & \boldmath$C$ &
\textbf{Tok time (s)} \\
\midrule
Fast        & C4        & 600  & 100    & 9.73   \\
Fast        & Wikipedia & 600  & 100    & 11.07  \\
Baseline    & C4        & 1000 & 2000   & 15.40  \\
High-acc    & C4        & 2000 & 10\,000 & 27.5  \\
\bottomrule
\end{tabular}
\caption{Runtime per token and per 10-token prompt (wall-clock seconds, single
NVIDIA RTX A6000).}
\label{tab:runtime}
\end{table}

\subsection{Comparison with the SIPIT Reference}
\label{subsec:ref-comp}

The point of comparing with the released SIPIT
reference~\cite{nikolaou2025languagemodelsinjectiveinvertible} is not to
establish that recovery is possible---that is already settled by the injectivity
result---but to locate the differences. The
reference \emph{hard-projects} the proxy onto a discrete token at every
optimisation step, bans that candidate, rotates to the next nearest token, and
exits the moment a projected token reproduces the target; its discrete decision
and its stopping criterion therefore coincide. The pipeline studied in this work does the opposite:
it keeps the proxy continuous throughout, commits once, and consequently leaves a
measurable post-commitment residual---the discrete loss that powers the failure
detector of Section~\ref{subsec:diagnostics}.

The two ends of this trade-off are quantified in
Tables~\ref{tab:ref-acc}--\ref{tab:ref-time}. Per-step projection like in SIPIT implementation is decisively
faster (0.65~s/token vs.\ 17.8~s Baseline and 30.6~s High-accuracy on the same
100 C4 prompts, i.e.\ \(\approx27\)--\(47\times\)) and reaches 100\% exact match,
because aggressive early exit avoids running the full step budget. The presented approach already reaches the correct neighbourhood in most cases
(Section~\ref{subsec:rank}), so the residual accuracy gap is a matter of how
often projection is checked, not of the optimisation landscape. 
This is genuine choice rather than a deficiency: hard projection buys speed and
peak accuracy but collapses the continuous and discrete views into one, whereas
keeping them separate is precisely what yields the rank trajectories, the
per-position curves, and---most importantly---the ground-truth-free failure
signal, which allow for this study of the embedding space. 

\begin{table}[t]
\centering

\begin{tabular}{lccrr}
\toprule
\textbf{Implementation}  & \textbf{EM (\%)} &
\boldmath$SM_{mean}$ \\
\midrule
Custom (Baseline)   & 73.0          & 0.929 \\
Custom (High-acc)  & \textbf{96.0}  & 0.990 \\
\texttt{sipit\_ref}   & \textbf{100.0} & \textbf{1.000} \\
\bottomrule
\end{tabular}
\caption{Accuracy comparison on the first 100 C4 prompts (seed 42, 10 tokens
each). }
\label{tab:ref-acc}
\end{table}

\begin{table}[t]
\centering

\begin{tabular}{lrr}
\toprule
\textbf{Implementation} & \textbf{Tok time (s)} \\
\midrule
Custom (Baseline) & 17.8  \\
Custom (High-acc) & 30.6  \\
\texttt{sipit\_ref}             & \textbf{0.65}  \\
\bottomrule
\end{tabular}
\caption{Runtime comparison on the first 100 C4 prompts (single NVIDIA RTX A6000, 10-token prompts).}
\label{tab:ref-time}
\end{table}

\subsection{Problematic Tokens}
\label{subsec:problematic}

\begin{figure}[t]
\centering
\includegraphics[width=\textwidth]{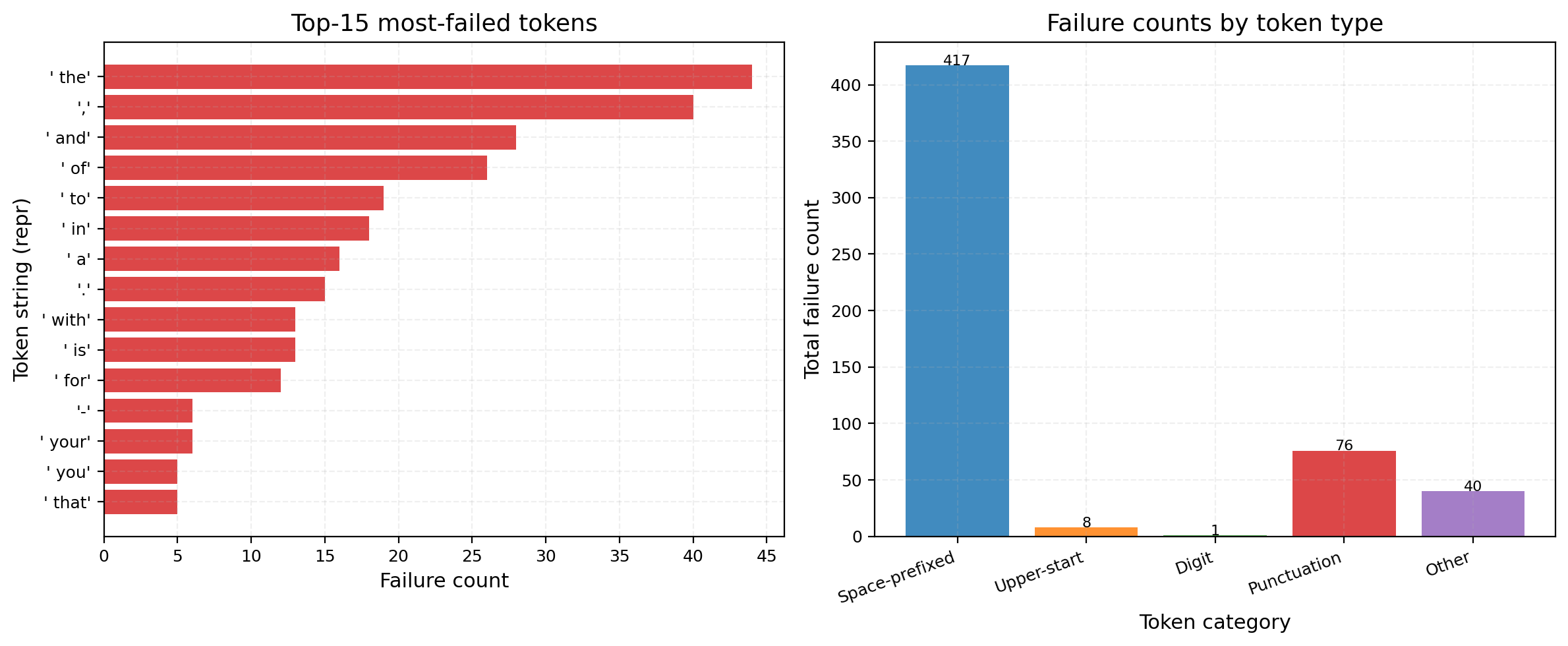}
\caption{Left: the 15 most frequently mis-committed ground-truth token strings
across the baseline batch (\(K=1000\), \(C=2000\), \(N=453\) C4 prompts). Right:
mis-committed tokens grouped by broad character-type category. Space-prefixed,
high-frequency function words dominate the failures, while content-bearing tokens
are almost never mis-committed.}
\label{fig:problematic}
\end{figure}

We analysed the frequency of token failures, depicted in Fig.~\ref{fig:problematic}.
Space-prefixed sub-words---those carrying the GPT-2 byte-level word-boundary
prefix that marks the start of a word---account for the overwhelming majority of
failures (417 of 542 across 450 prompts). The individual failure list is
dominated by the most common words in English:
`` the'', `` ,'', `` and'', and `` of''. This may be
counter-intuitive at first---the easiest tokens to guess are the hardest to
\emph{invert}---but it follows directly from embedding geometry. High-frequency
function words populate a dense region of the embedding matrix where many rows
are nearly collinear, so the proxy must converge with very high precision to make
the correct token the strict \(\ell_1\) nearest neighbour; a tiny residual is
enough to tip the commit onto a confusable neighbour. Punctuation forms the
second-largest category, while content-bearing tokens---proper nouns, technical
terms, digits---occupy well-separated regions and are recovered with better precision.

This asymmetry has a privacy implication. The exact-match rate is not a
uniform property of the text. This hints that the recovery of the less expected tokens (i.e. tokens containing random characters like random generated passwords), may be easier to recover than natural language.

\section{Discussion}
\label{sec:discussion}

The results of this study and the reconstruction of the embeddings in general bear directly on any system that transmits or stores last-layer
hidden states of a decoder-only model. In a \emph{split-inference} deployment
cut at the last layer, the leaked activation is exactly the object studied here:
an adversary with white-box access---including the server itself---can recover
the original query with high probability, and even the cheap Fast configuration
achieves non-trivial exact-match rates. For \emph{embedding APIs} that return
dense GPT-family vectors, the method applies directly, and the injectivity
result~\cite{nikolaou2025languagemodelsinjectiveinvertible} guarantees that the
mapping stays injective unless the architecture is fundamentally altered, even
under fine-tuning. The reported exact-match rates are \emph{lower bounds} on the
privacy risk: a high-similarity but non-exact recovery may still expose named
entities, medical terms, or financial figures, so even a similarity of 0.9 over
10 tokens represents near-complete exposure. Defences that restrict white-box
access partially mitigate the risk; differential privacy~\cite{dwork2014dp} or
post-hoc Gaussian noise analogous to DP-SGD~\cite{abadi2016dpsgd} applied to the
hidden states could reduce effectiveness at the cost of representation quality.
The overall recommendation is that last-layer hidden states be treated as
equivalent in sensitivity to the original text.

The gradient-based procedure is architecture-agnostic and requires only that the
model expose its hidden states and embedding matrix; larger GPT-family models
share the same causal architecture and injectivity conditions, so the main cost
at scale is the more expensive per-step forward pass. Whether the accuracy
ceiling shifts with size is open: richer embedding spaces may either help
(better-separated neighbourhoods) or hinder (denser clusters) the commit step.
Encoder-only (bidirectional) and encoder--decoder models fall outside the
decoder-only injectivity proof and remain an open question. Fine-tuning
preserves the architecture and, under the SIPIT conditions, injectivity, but
reshapes the embedding landscape; domain-specific fine-tuning may even ease
inversion by narrowing the effective vocabulary.

\section{Conclusion and Future Work}
\label{sec:conclusion}

We studied hidden-state inversion of GPT-2 small not as a one-shot
reconstruction but as a continuous, projection-free embedding-space optimisation,
and argued that this framing is where the interesting structure lives. Keeping
the proxy continuous throughout and committing a token only once makes the
attack \emph{observable}: it exposes rank trajectories, per-position convergence
curves, and a discrete loss at commit time. The most consequential of these is
the self-diagnostic signal---the cumulative discrete loss separates exact from
failed reconstructions by about ten orders of magnitude, letting an attacker or
auditor certify, with no ground-truth text, whether a given recovery succeeded.
The accompanying error analysis shows that failures are not uniform but
concentrate sharply on dense, high-frequency function-word tokens, so that
content-bearing, sensitive tokens are in fact the easiest to leak. Across
10-token C4 prompts the exact-match rate rises from 66.9\% to 97.5\% (mean
similarity 0.994) as the candidate window widens, confirming that most errors are
recoverable near-misses. A comparison with the released SIPIT reference is used
only to position this design: per-step hard projection is faster, but it
collapses the continuous and discrete views and forfeits the diagnostic residual
that our formulation provides.

The study is limited to GPT-2 small, to 10-token English sequences, and to a
white-box threat model, and it evaluates no defences. Future work includes
extending the analysis to larger and fine-tuned models and to longer or streaming
sequences (where the cascade mechanism is amplified), designing a sparser
projection schedule that closes the runtime gap \emph{while} retaining the
discrete-loss failure signal, and quantifying the privacy--utility trade-off of
differential-privacy and noise-injection defences---ideally guided by the same
discrete-loss diagnostic, which gives a direct read-out of how much a defence
degrades recoverability.

\begin{credits}
\subsubsection{\discintname}
The authors have no competing interests to declare that are relevant to the
content of this article.
\subsubsection{\ackname}
We gratefully acknowledge Polish high-performance computing infrastructure PLGrid (HPC Center: ACK Cyfronet AGH) for providing computer facilities and support within computational grant no. PLG/2025/018873
\end{credits}

\end{document}